\documentclass[10pt,twocolumn,letterpaper]{article}

\usepackage{cvpr}
\usepackage{times}
\usepackage{epsfig}
\usepackage{graphicx}
\usepackage{amsmath}
\usepackage{amssymb}
\usepackage{multirow}
\usepackage{color}
\usepackage{epstopdf}
\usepackage{bm}
\newcommand{\tabincell}[2]{\begin{tabular}{@{}#1@{}}#2\end{tabular}}

\usepackage[breaklinks=true,bookmarks=false]{hyperref}

\cvprfinalcopy 

\def\cvprPaperID{5862} 

\ifcvprfinal\fi

\begin{document}

\title{Connection Sensitive Attention U-NET for Accurate Retinal Vessel Segmentation}

\author{Ruirui Li, Mingming Li, Jiacheng Li\thanks{The corresponding author.},  Yating Zhou\\
Beijing University of Chemical Technology \\
North Third Ring Road 15, Chaoyang District\\
Beijing, China, 100029\\
{\tt\small \{ilydouble,lmming0429\}@gmail.com, 747530953@qq.com, m18810600393@163.com}
}

\maketitle

\begin{figure}

\centering
\includegraphics[width=0.8\linewidth]{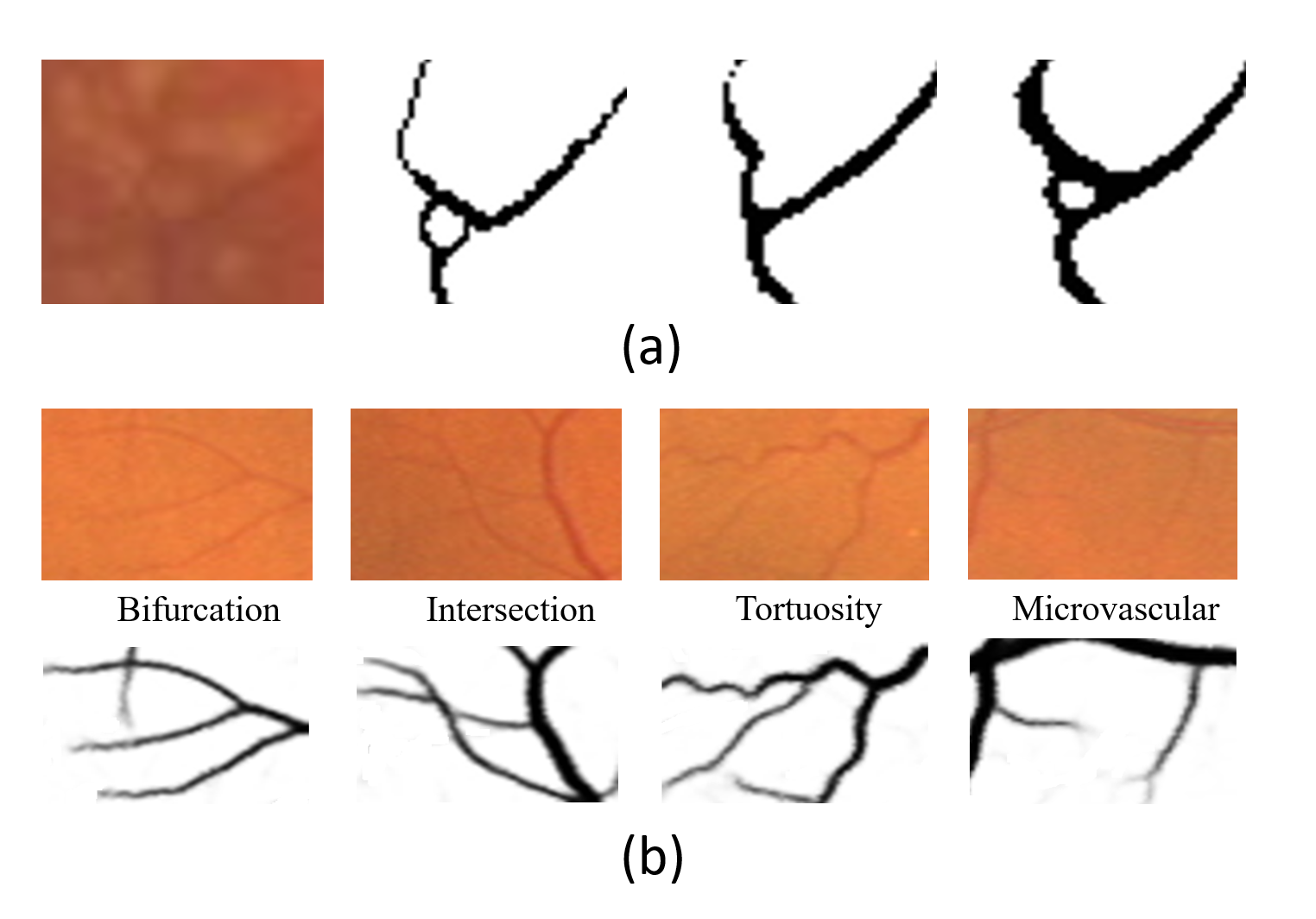}
\caption{Challenges on retinal vessel segmentation:(a) an example on STARE, from left to right: input, GT, VGAN and our method; (b) our results for bifurcation, intersection, tortuosity and microvascular cases.}
\label{fig:1}
\end{figure}

\begin{abstract}
We develop a connection sensitive attention U-Net(CSAU) for accurate retinal vessel segmentation. This method improves the recent attention U-Net for semantic segmentation with four key improvements: (1) connection sensitive loss that models the structure properties to improve the accuracy of pixel-wise segmentation; (2) attention gate with novel neural network structure and concatenating DOWN-Link to effectively learn better attention weights on fine vessels; (3) integration of connection sensitive loss and attention gate to further improve the accuracy on detailed vessels by additionally concatenating attention weights to features before output; (4) metrics of connection sensitive accuracy to reflect the segmentation performance on boundaries and thin vessels.

Our method can effectively improve state-of-the-art vessel segmentation methods that suffer from difficulties in presence of abnormalities, bifurcation and microvascular. This connection sensitive loss tightly integrates with the proposed attention U-Net to accurately (i) segment retinal vessels, and (ii) reserve the connectivity of thin vessels by modeling the structural properties. Our method achieves the leading position on DRIVE, STARE and HRF datasets among the state-of-the-art methods.

\end{abstract}

\section{Introduction}
\label{sec:introduction}
Retinal vasculature structure implicates important information and helps the ophthalmologist in detecting and diagnosing a variety of retinal pathology such as Retinopathy of Prematurity (RoP), Diabetic Retinopathy(DR), Glaucoma, hypertension, and Age-related Macular Degeneration(AMD) which are leading causes of blindness. The segmentation of retinal vessels is particularly important for diagnosis assistance, treatment and surgery planning of retinal diseases. Changes in vessel morphology such as shape, tortuosity, branching pattern and width provide an accurate early detection of many retinal diseases. 

Over the past two decades, a tremendous amount of research has been devoted in segmenting the vessels from retinal fundus images. Numerous fully automated methods\cite{vostatek2017performance,odstrcilik2013retinal,orlando2017retinal} have been proposed in literature which were quite successful in achieving segmentation accuracy on par with trained human annotators. Despite this, there is a considerable method for further improvements due to various challenges posed by the complex nature of vascular structures. Some of the active problems include segmentation in the presence of abnormalities, segmentation of thin vessels structures and segmentation near the bifurcation and crossover regions.

Comprehensive and detailed survey of retinal vessels segmentation methods are included in \cite{srinidhi2017recent, almotiri2018retinal, fraz2012blood}. Works that concerned by the paper are deep learning based methods for accurate retinal vessel segmentation. Liskowski et al. \cite{liskowski2016segmenting} proposed a deep neural network model, achieving an area under the curve (ROC AUC) of 0.97 on the DRIVE dataset. Their method performs reasonably well on pathological images. A novel CNN architecture was proposed in \cite{maninis2016deep} to solve both the retinal vessel and optic disc segmentation problem. Fu et al. \cite{fu2016retinal} formulated the vessel segmentation as a boundary detection problem using fully connected CNN model. In semantic segmentation field, U-Net\cite{ronneberger2015u} are fully convolutional networks for biomedical image segmentation.

Though many deep learning based approaches have been proposed, existing methods tend to miss fine vessels structures or allow false positives at terminal branches.Attention U-Net\cite{oktay2018attention} is used to automatically learn to focus on target structure of varying shapes and sizes.  Mosinska et.al \cite{mosinska2018beyond} have found that pixel-wise losses are unsuitable for retinal vessel segmentation because of their inability to reflect the topological impact of mistakes in the final prediction. The work\cite{wei2017road} added a coefficient to cross-entropy loss. It designed an estimating way of connectivity depending on the Euclidean distance between focused pixel and the nearest pixel belongs to the class. Ventura et.al\cite{ventura2017iterative} defined a new way to evaluate the connectivity on a patch. The most recent approach by Son et al. \cite{son2017retinal} generates the precise map of retinal vessels using generative adversarial training (GAN). Unfortunately, with limited data, generative models are considered much harder to train than discriminative models.

For thin vessels segmentation, this paper proposes an efficient topology-aware loss and a novel attention mechanism based on the U-Net to improve the accuracy. The proposed loss is called connection sensitive loss (CS loss) in that it considers the probability of connectivity in the neighboring region when designing the loss function. Moreover, the network is added new attention gates and learns a better matrix of attention weights before output. The proposed method provides an end-to-end fashion without any intervene in learning. With the well-designed attention U-Net architecture, the proposed connection sensitive loss gets the highest $F_1$-$score$ on all the three datasets which are DRIVE\cite{staal2004ridge}, STARE\cite{hoover2000locating} and HRF\cite{budai2013robust}. It also performs better to extract thin vessel structures compared with the state-of-the-art methods. In summary, the paper mainly made the following contributions:
\begin{enumerate}
\item For vessels segmentation, the paper proposes a connection sensitive loss. It is designed for simultaneous region-wise structure extraction and pixel-wise semantic segmentation. It helps achieve accurate results, even for thin vessel structures in crossover regions.
\item A new attention mechanism is designed based on the standard U-Net. The proposed attention gates improve the quality and the effectiveness of the features and thus take better advantage of them during segmentation.
\item The paper proposes the connection sensitive attention U-Net (CSAU) which combines the connection sensitive loss and the attention gates together. In the experiment, CSAU gets the highest $F_1$-$score$ on all the three datasets compared with the state-of-the-art methods.
\item In order to better reflect the quality of the segmentation details, this paper invents a new metrics to evaluate the segmentation of boundaries and thin vessel structures. We name it as connection sensitive accuracy.  
\end{enumerate}

In Section 2, we will introduce the proposed method. Section 3 shows implementation details that include data preprocessing and training process. And Section 4 discusses the experimental situation and analyzes the results. The last section shows the conclusions of this paper.
\section{Proposed methodology}
\label{sec:Proposed methodology}
\begin{figure}

\centering
\includegraphics[width=\linewidth]{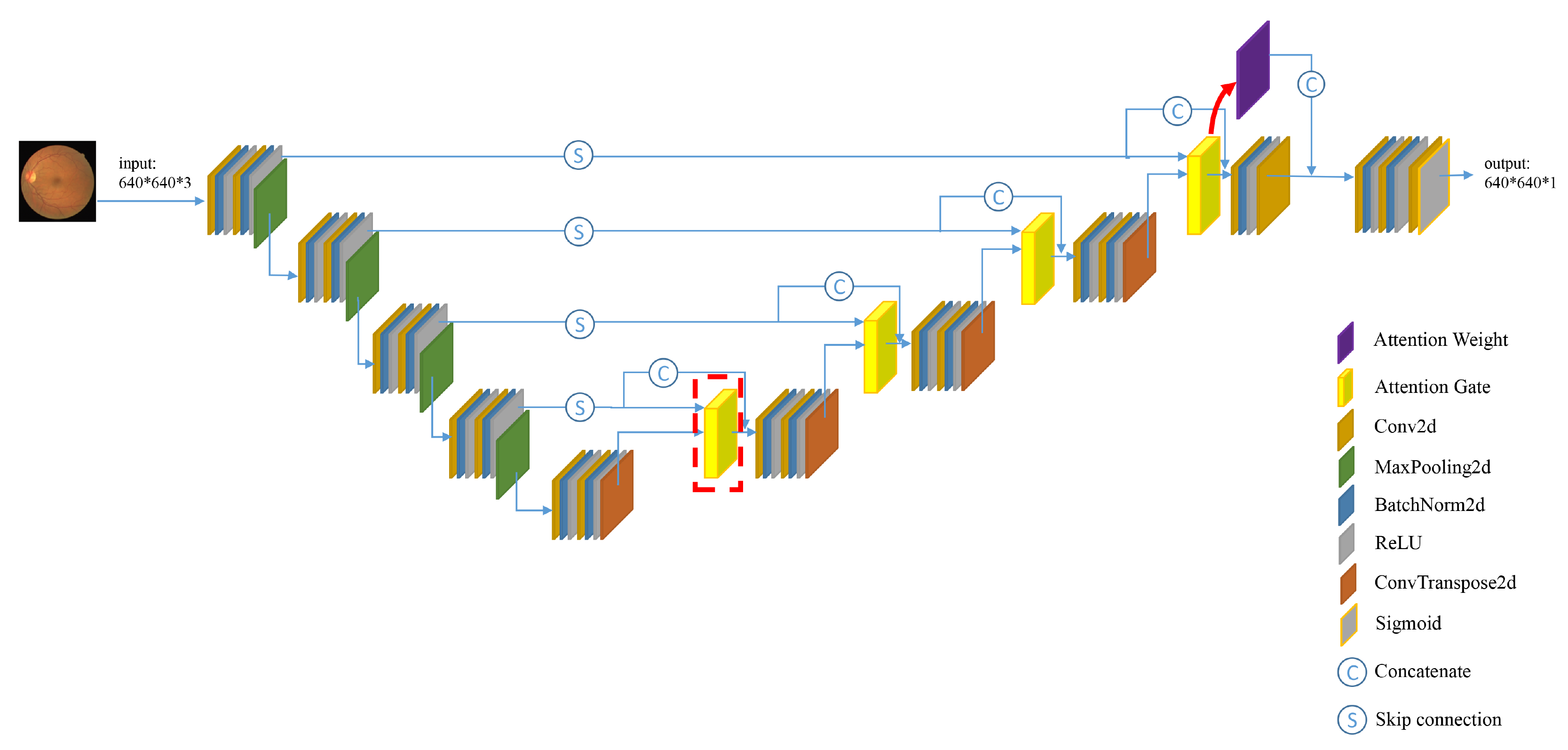}
\vspace{-5mm}
\caption{The proposed framework}
\label{fig:2}
\end{figure}

In this section, we present the architecture of the connection sensitive attention U-Net (CSAU). The main framework is showed in Fig. \ref{fig:2}. Its structure is very like the original attention U-Net except the connections and the designs of attention gates. Moreover, the framework uses a new connection sensitive loss with which the attention gate learns better attentive weights and helps improve the accuracy of details.

The parameters of the convolutional neural layers are listed in Table 1. The network contains four encoder blocks and four decoder blocks. They are connected by the skip connections. Each encoder block consists of two successive 3$\times$3 convolutional layers and a max pooling layer. Every convolutional layer is followed by a Batch-normalization layer and a ReLU layer. The decoder block is the same as the encoder block except that it uses the transposed convolutional layer instead of the pooling layer.

\begin{table}
\center
\caption{The parameters of the convolutional neural layers.}
\scalebox{0.8}{
\setlength{\tabcolsep}{1mm}{
\begin{tabular}{|c|c|c|c|}
\hline
\tabincell{c}{Block \\ name} & \tabincell{c}{Layer \\ name} & \tabincell{c}{Layer \\ configuration} & Remark \\
\hline
\multirow{2}{*}{\tabincell{c}{Encoder \\ Block(1)}} & conv1\_1 & 3$\times$3, 32 & \multirow{12}{*}{\tabincell{c}{Down-sampling \\ path}} \\
\cline{2-3} & conv1\_2 & 3$\times$3, 32 & \\
\cline{1-3}
\multicolumn{3}{|c|}{2$\times$2 max pool, stride 2} & \\
\cline{1-3}
\multirow{2}{*}{\tabincell{c}{Encoder \\ Block(2)}} & conv2\_1 & 3$\times$3, 64 & \\
\cline{2-3} & conv2\_2 & 3$\times$3, 64 & \\
\cline{1-3}
\multicolumn{3}{|c|}{2$\times$2 max pool, stride 2} & \\
\cline{1-3}
\multirow{2}{*}{\tabincell{c}{Encoder \\ Block(3)}} & conv3\_1 & 3$\times$3, 128 & \\
\cline{2-3} & conv3\_2 & 3$\times$3, 128 & \\
\cline{1-3}
\multicolumn{3}{|c|}{2$\times$2 max pool, stride 2} & \\
\cline{1-3}
\multirow{2}{*}{\tabincell{c}{Encoder \\ Block(4)}} & conv4\_1 & 3$\times$3, 256 & \\
\cline{2-3} & conv4\_2 & 3$\times$3, 256 & \\
\cline{1-3}
\multicolumn{3}{|c|}{2$\times$2 max pool, stride 2} & \\
\hline
\multirow{3}{*}{\tabincell{c}{Decoder \\ Block(5)}} & conv5\_1 & 3$\times$3, 512 & \multirow{12}{*}{\tabincell{c}{Up-sampling \\ path}} \\
\cline{2-3} & conv5\_2 & 3$\times$3, 512 & \\
\cline{2-3} & convTranspose5\_1 & 2$\times$2, 256 & \\
\cline{1-3}
\multirow{3}{*}{\tabincell{c}{Decoder \\ Block(6)}} & conv6\_1 & 3$\times$3, 256 & \\
\cline{2-3} & conv6\_2 & 3$\times$3, 256 & \\
\cline{2-3} & convTranspose6\_1 & 2$\times$2, 128 & \\
\cline{1-3}
\multirow{3}{*}{\tabincell{c}{Decoder \\ Block(7)}} & conv7\_1 & 3$\times$3, 128 & \\
\cline{2-3} & conv7\_2 & 3$\times$3, 128 & \\
\cline{2-3} & convTranspose7\_1 & 2$\times$2, 64 & \\
\cline{1-3}
\multirow{3}{*}{\tabincell{c}{Decoder \\ Block(8)}} & conv8\_1 & 3$\times$3, 64 & \\
\cline{2-3} & conv8\_2 & 3$\times$3, 64 & \\
\cline{2-3} & convTranspose8\_1 & 2$\times$2, 32 & \\
\hline
\multirow{2}{*}{} & conv9\_1 & 3$\times$3, 32 & \multirow{5}{*}{} \\
\cline{2-3} & conv9\_2 & 3$\times$3, 1 & \\
\cline{1-3}
\multirow{3}{*}{} & conv10\_1 & 3$\times$3, 32 & \\
\cline{2-3} & conv10\_2 & 3$\times$3, 32 & \\
\cline{2-3} & conv10\_3 & 3$\times$3, 1 & \\
\hline
\end{tabular}}}
\end{table}

\subsection{Connection sensitive loss}
\label{sec:csloss}

The parameters of the model are learnt by a training objective, using Adam stochastic gradient descent. In this paper, we build a new training objective on top of the proposed attention U-Net architecture. In the following discussion, let $x \in R^{H\times W}$ be the $H \times W$ input image, and let $y \in \{0, 1\}^{H\times W}$ be the corresponding ground-truth labeling, with 1 indicating pixels in the vessels and 0 indicating background pixels. Let $f$ be the proposed neural network  parameterized by weights $v$. The output of the network is an image $\hat{y} = f(x, v) \in \{0, 1\}^{H\times W}$. Every element of $\hat{y}$ is interpreted as the probability of pixel $i$ having label 1: $\hat{y}_i \equiv p(Y_i=1|x,v)$, where $Y_i$ is a random Bernoulli variable $Y_i \sim Ber(\hat{y}_i)$.

Cross entropy is widely used as the loss function in deep learning networks to deal with binary classification problems, which calculates the probability of being one specific class or not. Thus, the proposed loss function is also on the basis of the cross-entropy loss $L_{ce}$ defined by 
   
\begin{equation}
L_{ce} = -\sum_{i=1}(y_i * log(f_i(x,v))+(1-y_i) * log(1-f_i(x,v))
\label{eq:r1}
\end{equation}

\begin{figure}
\centering
\includegraphics[width=0.7\linewidth]{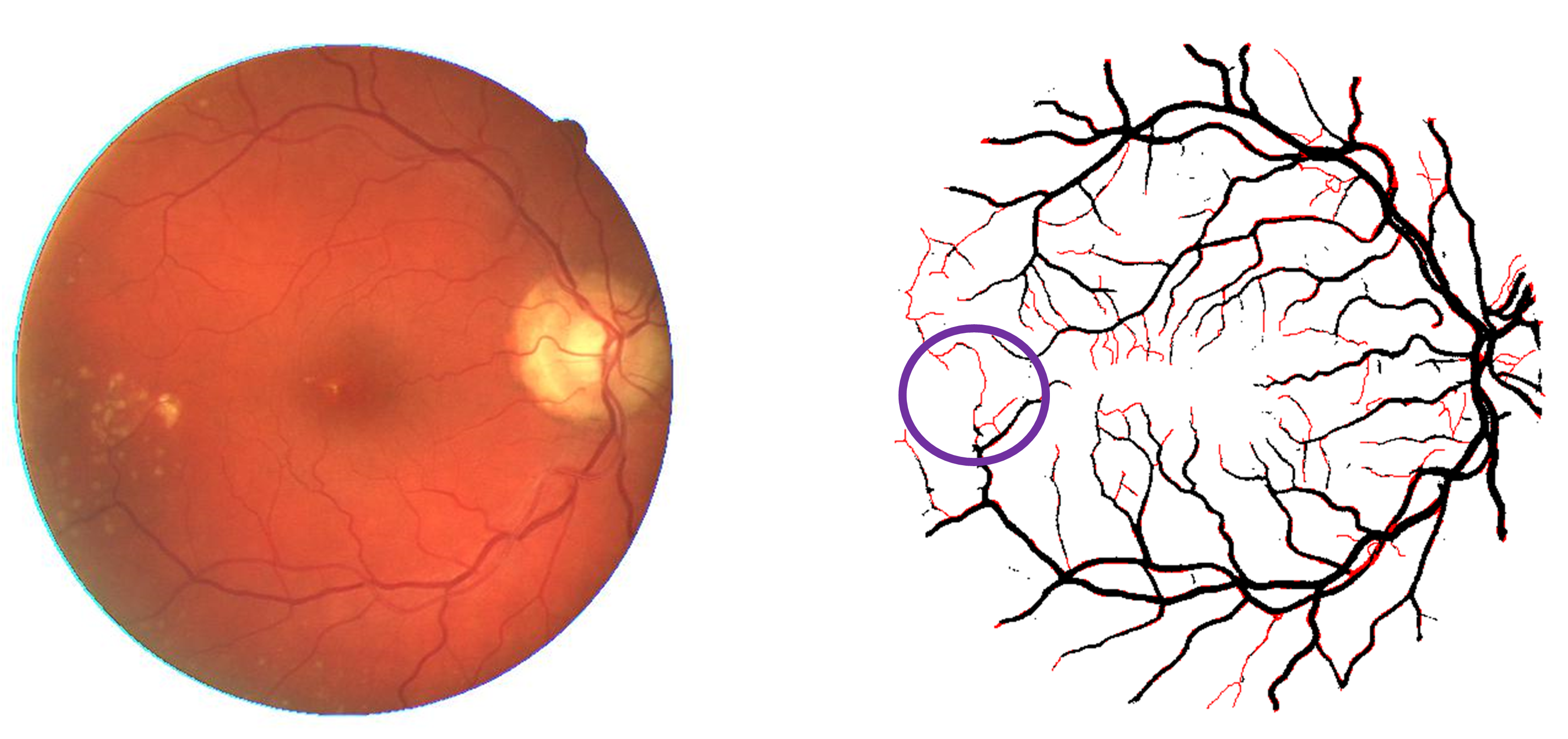}
\caption{Results trained by binary cross entropy in which red pixels are false negatives.}
\label{fig:3}
\end{figure}

By observing the definition of $L_{ce}$ in (\ref{eq:r1}), we can find that the cross-entropy loss assigns equal weights to the loss of different pixels, failing to consider fine object structures. Therefore, cross entropy loss is not fit well to the tasks of segmenting connected vascular structures. Fig. \ref{fig:3} shows the segmentation results produced by the U-Net with the cross-entropy loss. The colored pixels are false negative results. It is obvious that cross-entropy loss tends to bring broken vessels in terminal branches, which are critical for diagnosis.

The connection sensitive loss is designed for neural network training tasks in the field where the structural connectivity of segmented objects is concerned. To solve the problem, we take the connectivity into consideration by encoding two coefficients into the cross-entropy loss, as showed in (3). $L_{cs}$ is the connection sensitive loss. $\theta_1$ and $\theta_2$ represent local structural properties in the labeled ground truth and the predicted map respectively, while $w_i$ is a weighted parameter that multiplies with the encoded loss on every pixel which will be explained later. 
\begin{equation}
L_{cs}\!=\!-\!\sum_{i=1}\!w_i * (\theta_1 y_i log(f_i(x,\!v)\!)+\theta_2 (1-y_i)log(1-\!f_i(x,\!v)\!)
\label{eq:r2}
\end{equation}

To model the structural properties, an exponential function is constructed as showed in the following equation:
\begin{equation}
\theta_1 = e^{(1-{C_i}^2*y_i)}, \theta_2 = e^{(1-{C_i}^2*f_i(x,v))}   \label{eq:r3}
\end{equation}

in which $C_i$ represents the probability of connectivity in local regions. It can be computed by the following function with upper bound 1 and lower bound 0. $z_i$ is a variable representing whether the pixel belongs to the ground truth $(z_i = y_i)$ or the predicted map $(z_i = f_i(x,v))$.
\begin{equation}
C_{i,i \in \Omega (m,n,r;z)}\!=\! max(\!min(\!\alpha * (\!\frac{\sum_{i \in \Omega (m,n,r;z)}}{r^2})^\beta - \gamma,\! 1),\! 0)
\label{eq:r4}
\end{equation}

It is observed that $C_i$ is strongly correlated to the local density. To estimate $C_i$, the function chooses a polynomial model and computes the local density by averaging the values in the region. $\alpha, \beta, \gamma$ are constant coefficients. $\Omega (m,n,r;z)$ represents a square region in the map $z$ with the side length $r$ and the coordinate $(m,n)$ of the center point. The region can be defined with the matrix in the equation:
\begin{equation}
\Omega_{(\!m\!,n\!,r;z\!)\!}\!=\!\left[ \! \begin{matrix}
Z_{(m\!-\!\frac{r-1}{2},n\!-\!\frac{r-1}{2})}\! & \cdots\! & \cdots\! & Z_{(m\!-\!\frac{r-1}{2},n+\frac{r-1}{2})} \\
\cdots\! & \cdots\! & \cdots\! & \cdots \\
\cdots\! & \cdots\! & \cdots\! & \cdots \\
Z_{(m\!+\!\frac{r-1}{2},n\!-\!\frac{r-1}{2})}\! & \cdots\! & \cdots\! & Z_{(m\!+\!\frac{r-1}{2},n\!+\!\frac{r-1}{2})}
\end{matrix}
\!\right] _{r\!\times \! r}
\end{equation}

\begin{figure}
\centering
\includegraphics[width=0.6\linewidth]{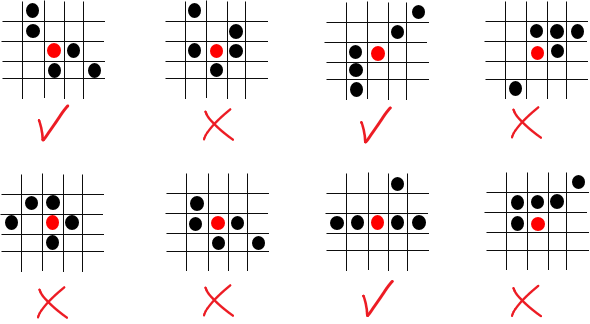}
\caption{Some samples when adding 5 points in 5$\times$5 resolution area.}
\label{fig:4}
\end{figure}

\begin{figure}
\centering
\includegraphics[width=0.6\linewidth]{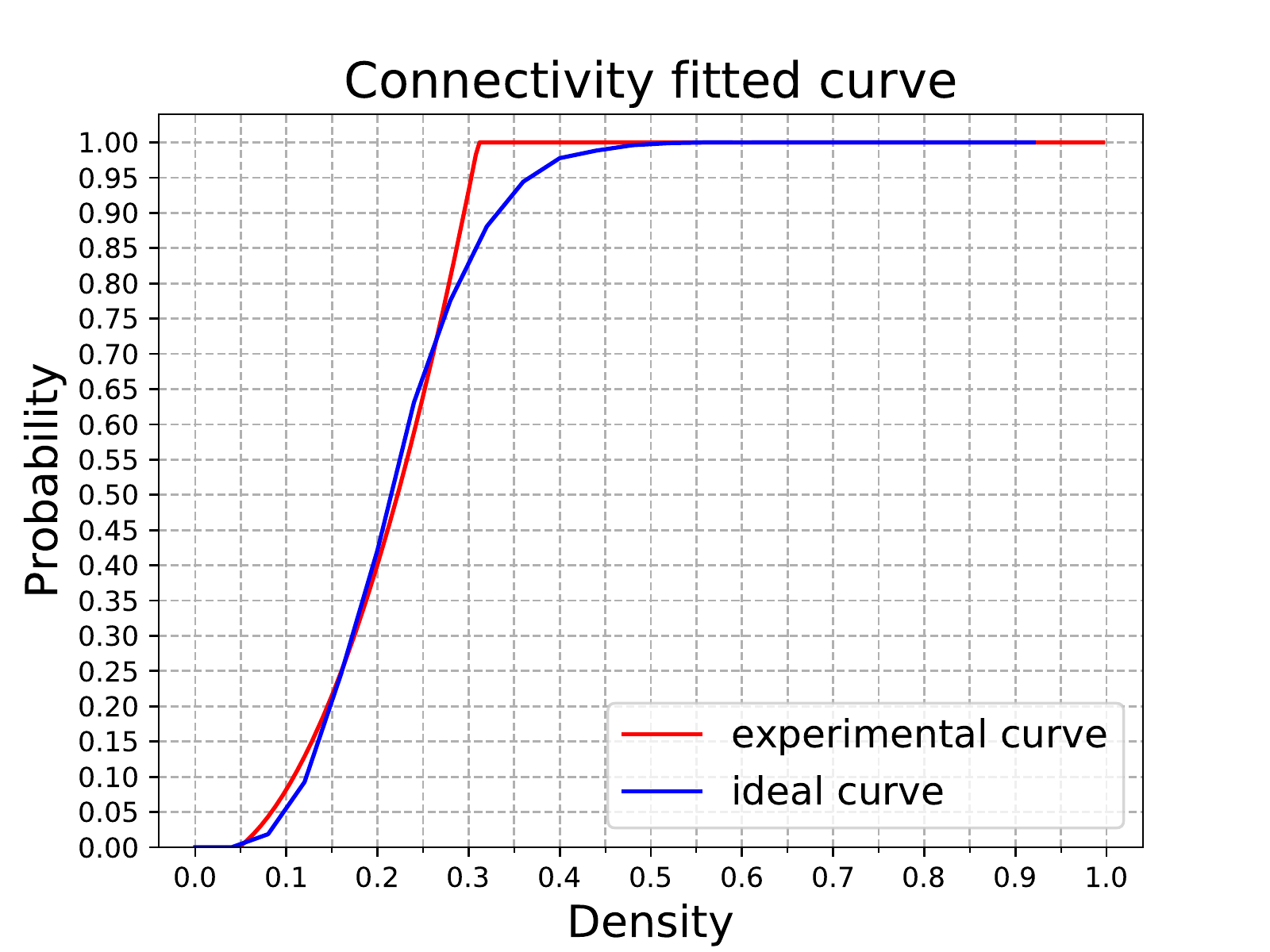}
\caption{Curves of connectivity probability with different densities in 5$\times$5 region.}
\label{fig:5}
\end{figure}

  To get the values of the constant coefficients $\alpha, \beta$ and $\gamma$, we throw N sampling points on $r \times r$ region for different densities through the Monte Carlo important sampling. Inspired by the definition of connectivity in the paper\cite{ventura2017iterative}, on each sampled patch, we decide whether the region is connected or not by checking if there exist two paths from the center point $(m, n)$ to the boundary of the region according to the eight-connected domain algorithm. Fig. \ref{fig:4} shows some cases when the density is 0.2 in $5\times 5$ resolution areas. Fig. \ref{fig:5} shows the fitted curve when $\alpha = 10.3180$, $\beta = 1.9808$, $\gamma = -0.0254$ and $r=5$. The sampled blue curve is very close to the modeled red curve.
  
\begin{figure}
\centering
\includegraphics[width=\linewidth]{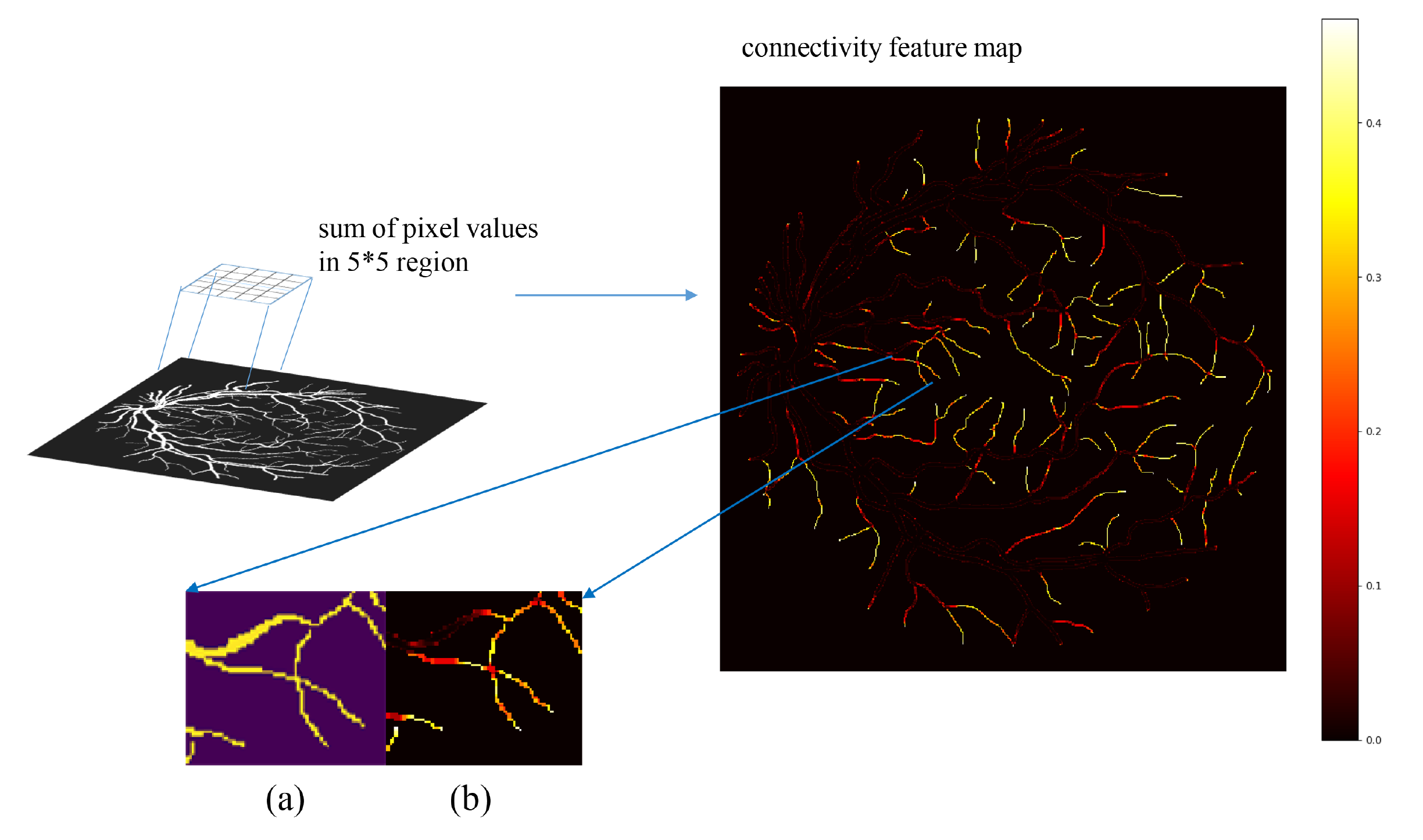}
\caption{(a) is a local region of the label image, the purple region is the background and the yellow region is the vessels. (b) is the corresponding region of connectivity feature map where the pixels with dark value have higher probability of connectivity than those with bright colors.}
\label{fig:6}
\end{figure}

It is recommended to choose $r=5$ during the local connectivity estimation for simplification without scarifying too much accuracy. In fact, $5\times5$ area could be seen as a local pattern. Images with complex contents and other resolutions could be mapped to the local pattern.
Fig. \ref{fig:6} illustrates the connectivity feature map in which pixels are computed through $(y_i - C^{2}_{i} \times y_i)$ on the ground truth. It means that the larger the value is, the more attention should be paid on this pixel. It is assumed that large value has high risk of being less connected.

The factor $w_i$ is proposed to further decrease the false negatives and is formulated as:
\begin{equation}
w_i = (1+(max(\Omega_{j,i,\lambda;o}) - f_i(x,v))*y_i)
\end{equation}

If the output $f_i(x,v)$ is expected to connect other vessel pixels and is predicted small probability, the value $w_i$ would be higher and brings more punishment on the false negative pixels by increasing their losses. The punishment is region-aware. The term $max(\Omega_{j,i,\lambda;o})$ indicates the probability to classify the pixel as vessel class to some extent. The larger the value is, the easier it is going to be recognized, and vice versa. The term $(max(\Omega_{j,i,\lambda;o}) - f_i(x,v))$ illustrates the difference between the probability and the predicted value. It is expected to become smaller during the training process.

\subsection{Attention gates}
\label{sec:ag}
The proposed attention gates are incorporated into the standard U-Net architecture to highlight salient features that are passed through the skip connections, see Fig. \ref{fig:2}. The attention gate has two input signals. One is the feature map that is transported by the skip connection. The other input is the coarse feature gotten from the output of  previous neural layer. Information extracted from coarse scale is used in gating to disambiguate irrelevant and noisy responses in skip connections. The output of attention gate is connected to the next decoder. The gating signal for each skip connection aggregates information from multiple imaging scales which increases the resolution of the attention weights and helps achieve better performance.
 
\begin{figure}
\centering
\includegraphics[width=\linewidth]{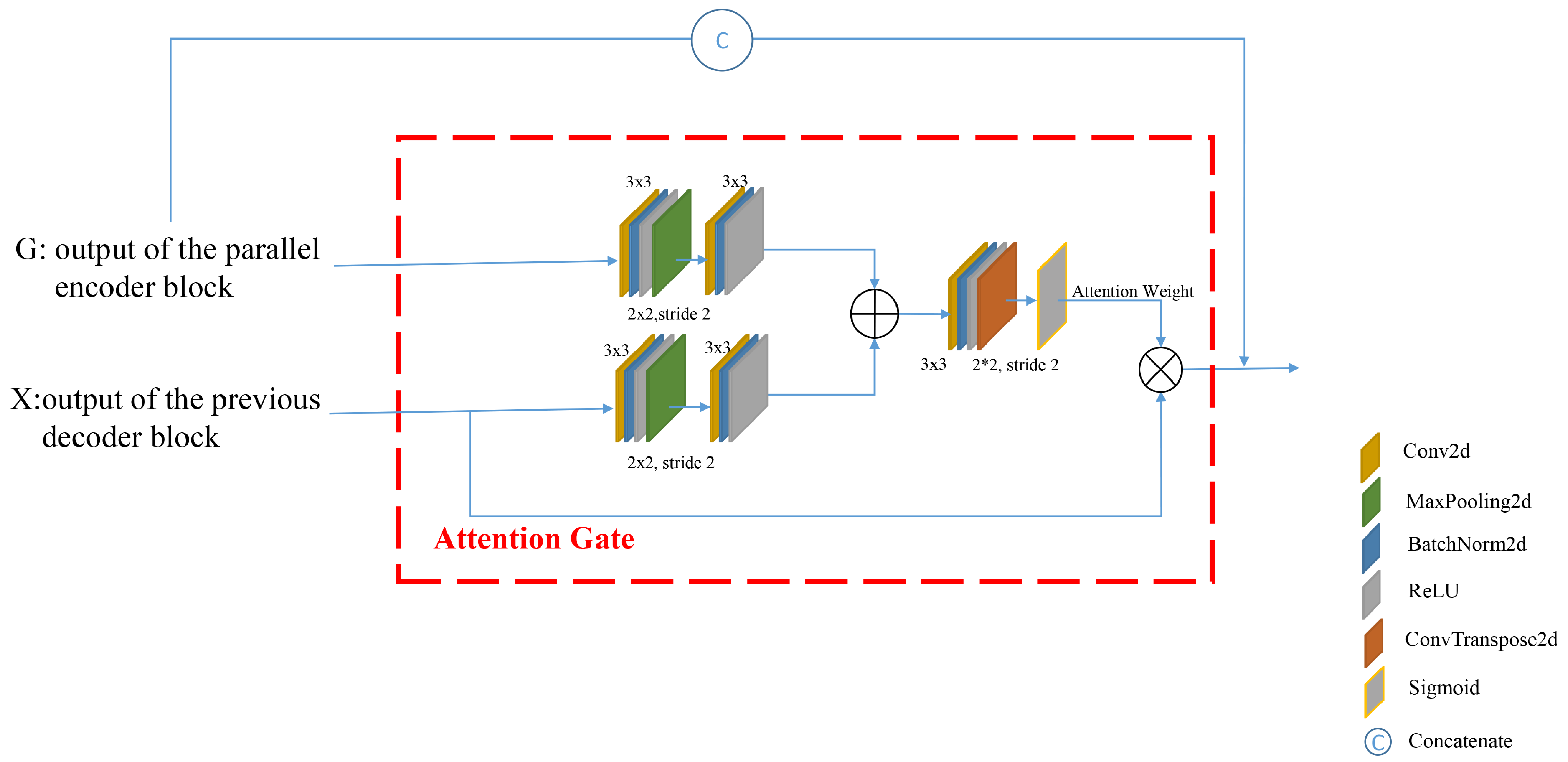}
\vspace{-5mm}
\caption{The proposed attention gate.}
\label{fig:7}
\end{figure}

The proposed attention gate is showed in Fig. \ref{fig:7}. It is actually a sub-network in a simple encoder-decoder pattern. The attention gate consists of five $3\times3$ convolutional layers, five batch normalizers, five ReLUs, two max pooling layers and a transposed convolutional layer. The feature map X and G are transformed to an intermedia space first. Then the addition of them are up-sampled by transposed convolution. We use additive attention\cite{bahdanau2014neural} to obtain the gating coefficient. Additive attention is formulated as follows:
 
\begin{equation}
\alpha_i = \sigma(q_{att}(x_i, g_i; \theta_{att}))
\end{equation}

where $\sigma(x_i) = \frac{1}{1+exp(-x_i)}$ correspond to sigmoid activation function. Attention gate is characterized by a set of parameters $\theta_{att}$ containing: linear transformations, non-linear transformations and bias terms. $q_{att}$ defines the operations on $x_i$ and $g_i$ by parameters $\theta_{att}$.

\begin{figure}
\centering
\includegraphics[width=0.7\linewidth]{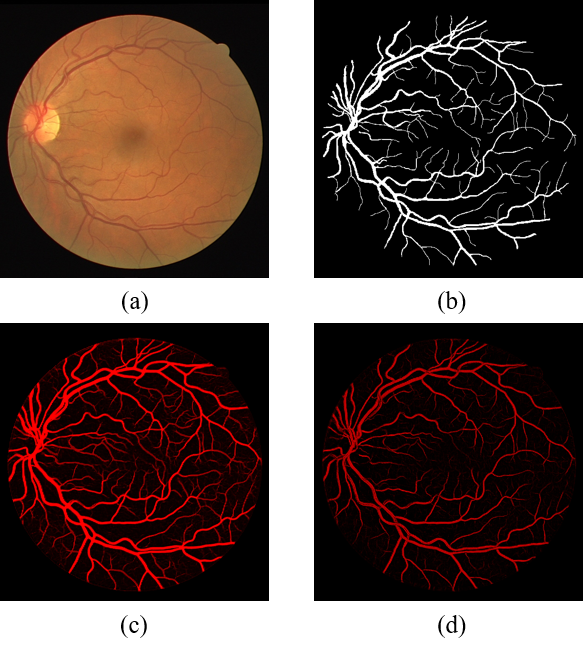}
\caption{Visualizations of attention weights. (a) is a fundus image, (b) is the ground truth and (c) (d) are visualized attention weights by UP-Link and DOWN-Link respectively.}
\label{fig:8}
\end{figure}

We tried two kinds of connection modes for designing of attention gates. We called them the UP-Link and the DOWN-Link respectively. According to the UP-Link, there is a connection between the input G and the output of attention gate as showed in Fig. \ref{fig:7}. On the other hand, the DOWN-Link has a connection between the input X and the output instead. CSAU chooses the UP-Link mode since such mechanism improves the quality and influence of detailed features during training. Updating parameters of the attention gates depends on the gradient passed not only from the decoder layers but also from the encoder layers. It results experimentally in better attention weights for segmentation model. Examples of intermediate attention weights are converted and visualized in Fig. \ref{fig:8} in which (c) illustrates the last attention weights gotten by the UP-Link while (d) illustrates that gotten by the DOWN-Link in the same situation. The UP-Link mode provides sufficient detailed information as well as strengthened salient features for the following decoders in feed forward propagation. As a result, both the vessels and the structures are well preserved. 

At the end of the network showed in Fig. \ref{fig:2}, the last attentive weights are extracted out and concatenated to the output of the features, which further emphasize attentive pixels. Experiments in section 4.3 show the validation of the proposed attention mechanism for thin vessels segmentation and its connectivity preservation.

\subsection{Metrics of connection sensitive accuracy}
General metrics for image segmentation could judge how good main vessels are segmented. But they could not distinct clearly the minor changes in boundaries and fine vessel structures which are critical for early diagnosis. To solve the problem, this paper presents a new evaluation metrics to evaluate the performance of segmentation on boundaries and thin structures. Based on a factor of CS loss, we define the $ACC_{cs}$ as follows:
\begin{small}
\begin{equation}
ACC_{cs} = \frac{\sum (m_i \times \delta_{1} (f_i (x,v)))}{\sum m_i}
\end{equation}
\end{small}

\begin{figure}
\centering
\includegraphics[width=0.5\linewidth]{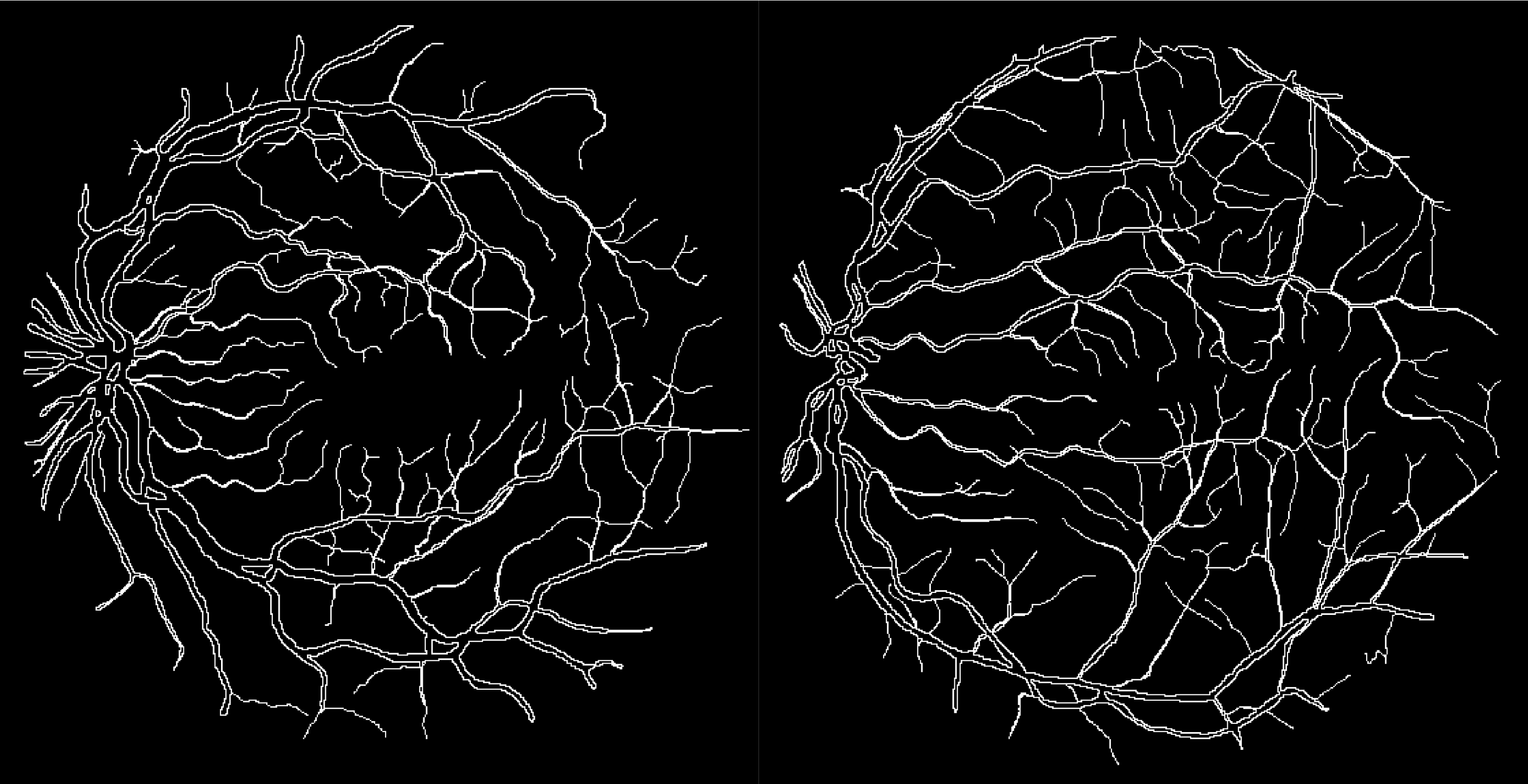}
\caption{Two examples of mask maps on DRIVE.}
\label{fig:m}
\end{figure}

\begin{small}
\begin{equation}
m_{i} = (\delta_{2}(1 - C^{2}_{i}) \times y_{i}) \cup DOG(y)
\end{equation}
\end{small}

in which $\delta_1$, $\delta_2$ are binary threshold functions. $f_{i}$ is the predicted result with input $x$ and weights $v$. $m_{i}$ constructed a mask map. It is calculated by union operation of two sets. The first set ($\delta_{2}((1 - C^{2}_{i}) \times y_{i}$) represents the pixels belonging to fine vessel structures that are hard to be segmented. The second set is the extracted boundary of the ground truth through the DOG edge detection algorithm. Fig. \ref{fig:m} presents two examples of the mask maps on No.3 image and No.11 image of DRIVE. Actually, $ACC_{cs}$ computes the proportion of correctly segmented pixels and the total pixels with the mask.

\section{Implementation and Experiments setup}
\label{sec:Implementation}

\subsection{Implementation details}
\label{sec:implementation details}
In this part, we will make a brief introduction of the implementation of the connection sensitive attention U-Net. The experiments are carried out on a laboratory computer. Its configuration is showed in Table 2. The operating system is Ubuntu 16.04. The main required packages include python 3.6, CUDA8.0, cudnn7.0, Pytorch0.4.0.

\begin{table}[htb]
\caption{Experimental environments}
\centering
\scalebox{0.8}{
\begin{tabular}{c|c}
\hline
\textbf{CPU} & Intel (R) Core (TM) i7-4790K 4.00Hz \\
\hline
\textbf{GPU} & GeForce GTX1080 Ti \\
\hline
\textbf{RAM} & 20GB \\
\hline
\textbf{Hard disk} & Toshiba SSD 512G \\
\hline
\textbf{System} & Ubuntu 16.04 \\
\hline
\end{tabular}}
\end{table}

To avoid complex CUDA coding, we make full use of functions provided by PyTorch, mainly the nn.Functional.conv2d and the nn.MaxPool2d. Specifically, to calculate the summation of the probability in the region that centered at a focused pixel, we use nn.Functional.conv2d with a kernel 5$\times$5 and perform convolution on the whole image, except the padding part, which remains zero. To get the max probability of  that region, we use nn.MaxPool2d, setting kernel size as 7.

\subsection{Datasets and preparation}
\label{sec:Datasets}
Our approach is examined on three widely used benchmarks: DRIVE\cite{staal2004ridge}, STARE\cite{hoover2000locating} and HRF\cite{budai2013robust}, provided by different organizations. All photographs in these benchmarks are RGB images, while annotated images are binary images. DRIVE contains 20 training images and 20 testing images, with each of size 584$\times$565. STARE contain 20 fundus images, with each of size 605$\times$700. We manually divide the STARE dataset into training and testing images in the ratio of 10/10. For DRIVE and STARE, we use only one image from the  training set for validation. The HRF dataset comprises 45 images and is organized as 15 subsets. Each subset contains one healthy fundus image, one image of patient with diabetic retinopathy and one glaucoma image. We set the first 5 subsets as our training set and the rest as testing set. Five validation images are randomly selected in the training set. 

For DRIVE, we resize each image to 640$\times$640 by padding it with zero in four margins. For STARE, we resize them to 720$\times$720 in the same way. Each image in HRF is digitalized to 2336$\times$3504 pixels. Because of the high resolution image in HRF and limitation of GPU memory, we crop a single image into 640$\times$640 tiles, and test the tiles one by one from bottom left to up right in a sliding window way. To predict the pixels in the border region of the image, the missing context is extrapolated by mirroring the input image. We use an overlap strategy  described in the work\cite{Li2017DeepUNet}. For each tile, we compute the weight for overlapped pixels by the Gaussian function.  Through weighted summary, we composite the overlapped tiles and seamless stitch the whole segmental image. 

To augment the data, the method rotates the image every 4 degree along the whole round. Then it further flips them horizontally and vertically. Thus, there are 270 images generated from a single image. 

\subsection{Training methodology}
\label{sec:Training methodology}
The model is trained by AdamW\cite{loshchilov2017fixing} with parameters $\beta_1 =0.9$, $\beta_2 =0.999$ and learning rate 0.002. We propose a new learning strategy for the experiments. According to the strategy, we test the latest model on the validation set for every fifty batches. We use its loss as metrics to adjust the following learning rate. If the loss doesn't decrease for continuous five groups of validations, the learning rate will be set to the maximum of the values between 0.0001 and 0.1 times current learning rate. If the loss doesn't decrease for continuous 20 groups of validations, the learning rate will be set to the initial value 0.002.

We use a mini-batch size of 2 images for DRIVE, STARE and HRF. The model with the minimal validation loss will be chosen as the final model for testing. According to the experiments, the validation loss tends to converge within 20$th$ training epoch and we set the max training epoch to 25.
\section{Results and Analysis}
\label{sec:Results and Analysis}
\subsection{Evaluation metrics}
\label{sec:Evaluation metrics}
We use $F_1$-$score$, PR AUC, ROC AUC, $Accuracy$ and $Sensitivity$ to evaluate the performance of binary segmentation model. False Negative($FN$), True Positive($TP$), True Negative($TN$), False Positive($FP$) are four basic elements to compute the metrics. We also introduce connection sensitive accuracy($ACC_{cs}$) to measure the performance of segmentation on terminal thin vessels.

\bm{$F_1-Score$} considers both Recall and Precision, which is defined as:

\begin{small}
\begin{equation}
Recall = Sensitivity = \frac{TP}{TP+FN}
\end{equation}
\end{small}

\begin{small}
\begin{equation}
Precision = \frac{TP}{TP+FP}
\end{equation}
\end{small}

\begin{small}
\begin{equation}
F_1 = 2 * \frac{Precision*Recall}{Precision+Recall}
\end{equation}
\end{small}

$F_1$-$Score$ is positively related to the performance of the model.

\bm{$Accuracy$} is the proportion of the pixels which are correctly segmented and the total pixels.

\begin{small}
\begin{equation}
Accuracy = \frac{TP+TN}{TP+FN+TN+FP}
\end{equation}
\end{small}

\textbf{PR AUC} and \textbf{ROC AUC} A $Precision$ and $Recall$ (PR) curve is plotting Precision against Recall while a Receiver Operating Characteristic (ROC) curve is plotting True Positive Rate ($Recall$) against False Positive Rate ($FPR$). $FPR$ is defined as:

\begin{small}
\begin{equation}
FPR = \frac{FP}{FP+TN}
\end{equation}
\end{small}

AUC is the area under the curve and the performance of the model is positively related to the value of the area.

\subsection{Overall performance}
\label{sec:Overall performance}
We trained the CSAU model on DRIVE, STARE and HRF respectively and compared it with the state-of-the-art methods. The results for comparison on DRIVE and STARE are obtained from the web site of VGAN\cite{son2017retinal}. We directly use the segmented images to compute the metrics. On the other hand, the results for comparison on HRF are gotten from the work\cite{orlando2017retinal}. Since they do not provide the source code and the result images, as a result, we simply copy the metrics provided in their paper. To guarantee the fairness, we use the same way when choosing training, validating and testing set, which is described in section 3.2.

\begin{figure}
\centering
\includegraphics[width=\linewidth]{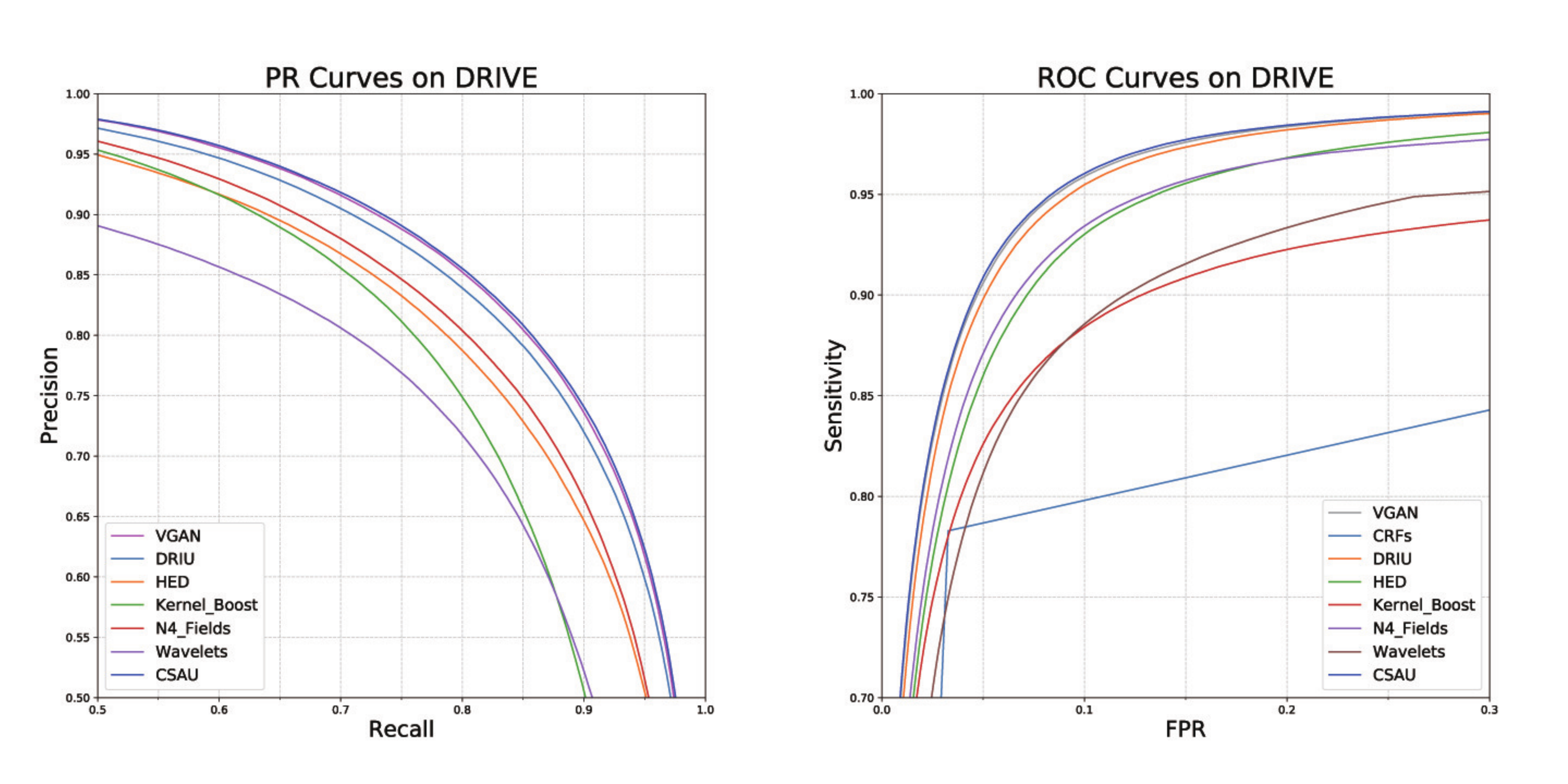}
\vspace{-5mm}
\caption{Precision and Recall curves and Receiver Operating Characteristic curves for different methods on DRIVE.}
\label{fig:9}
\end{figure}

\begin{figure}
\centering
\includegraphics[width=0.7\linewidth]{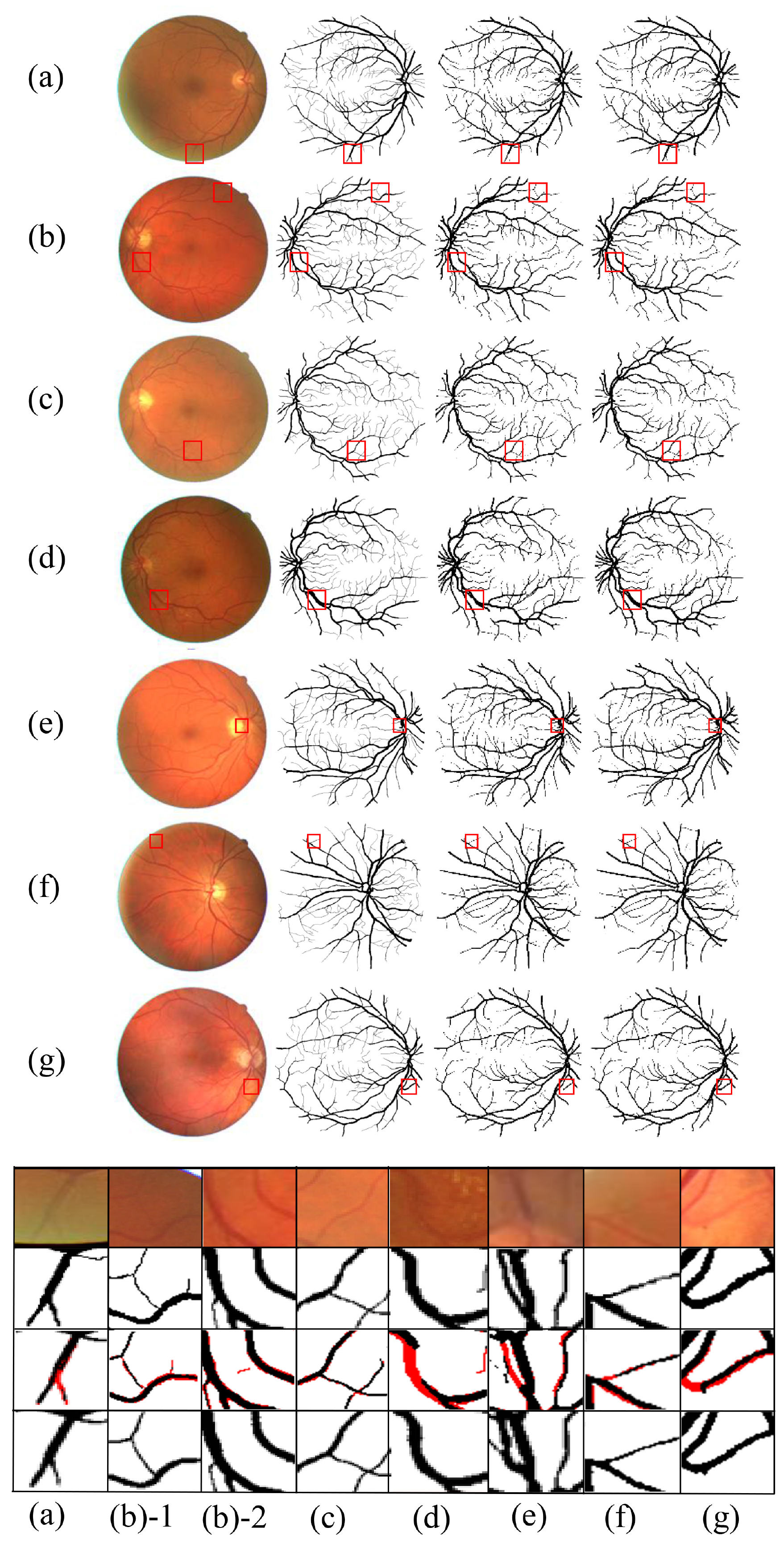}
\caption{Comparison of details between VGAN and CSAU.}
\label{fig:10}
\end{figure}

Table 3-5 show the results of comparison metrics. As observed, CSAU got the highest $F_1$-$score$, $Sensitivity$ and ROC AUC on all the benchmarks. On DRIVE, the proposed method achieves leading position on the leading broad through all the evaluation metrics. The comparison methods include K-Boost\cite{becker2013supervised}, HED\cite{xie2015holistically}, Wavelets\cite{soares2006retinal}, $N^{4}$-Fields\cite{ganin2014n}, DRIU\cite{maninis2016deep}, CRFs\cite{orlando2014learning} and VGAN\cite{son2017retinal}. Among them, HED, DRIU and VGAN are deep learning based methods which show superior performance in contrast to the other non-deep learning methods. Fig. \ref{fig:9} displays the PR curves and the ROC curves.  The performance of VGAN is also good and is listed in the second place. Compared with VGAN, the $F_1$-$score$ of CSAU is 0.2\% higher than that of VGAN and the $Sensitivity$ of CSAU is promoted by 0.6\%. CSAU improves the PR AUC by 0.2\% and ROC AUC by 0.4\% respectively. Actually, most deep learning based neural networks could segment the main vessels well. What really challenging is the task to segment thin vessel structures. In fundus images, pixels of thin vessels take a much smaller proportion compared with the other pixels. As a result, even the improvement on thin vessel segmentation is obvious, the promotion is slight when evaluated by the general metrics on the whole image. The last column in Table 3 shows the results of connection sensitive accuracy metrics by different methods. The $ACC_{cs}$ of CSAU is 2.8\% higher than that of VGAN and is 3.8\% higher than that of DRIU. It means that CSAU has a better performance on segmenting the boundaries and the thin vessels. Fig. \ref{fig:10} shows a group of examples on DRIVE. The segmented results by VGAN and CSAU are looked similar from an overall perspective. By zooming in the area surrounded by red rectangles, it is clear to distinct that where VGAN tend to obtain inaccurate boundaries and broken thin vessels. CSAU gets more accurate boundary and more integrated vessel structures.

\begin{table}
\center
\caption{Comparison of different methods on DRIVE.}
\scalebox{0.75}{
\setlength{\tabcolsep}{0.7mm}{
\begin{tabular}{ccccccc}
\hline
\multirow{2}{*}{Methods} & \multicolumn{6}{c}{DRIVE} \\
\cline{2-7} & ROC AUC & PR AUC & $F_1$-$score$ & $Sensitivity$ & $Accuracy$ & $ACC_{cs}$ \\
\hline
K-Boost\cite{becker2013supervised} & 0.9307 & 0.8464 & 0.7797 & 0.7563 & 0.9456 & 0.6739 \\
\hline
HED\cite{xie2015holistically} & 0.9696 & 0.8773 & 0.7938 & 0.7943 & 0.9475 & 0.7016 \\
\hline
Wavelets\cite{soares2006retinal} & 0.9436 & 0.8149 & 0.7601 & 0.7628 & 0.9387 & 0.6839 \\
\hline
$N^{4}$-Fields\cite{ganin2014n} & 0.9686 & 0.8851 & 0.8021 & 0.7994 & 0.9498 & 0.7178 \\
\hline
DRIU\cite{maninis2016deep} & 0.9793 & 0.9064 & 0.8210 & 0.8261 & 0.9541 & 0.7470 \\
\hline
CRFs\cite{orlando2014learning} & -- & -- & 0.7799 & 0.7829 & 0.9438 & 0.6785 \\
\hline
VGAN\cite{son2017retinal} & {\color{blue}0.9803} & {\color{blue}0.9142} & {\color{blue}0.8277} & {\color{blue}0.8300} & {\color{blue}0.9560} & {\color{blue}0.7537} \\
\hline
CSAU & {\color{red} 0.9807} & {\color{red}0.9157} & {\color{red}0.8294} & {\color{red}0.8349} & {\color{red}0.9563} & {\color{red}0.7751} \\
\hline
\end{tabular}}}
\end{table}

On STARE, similar phenomenon could be found as that on DRIVE in the experiments. CSAU wins the first place by all the metrics except the PR AUC. It gets 0.34\% higher ROC AUC, 0.1\% higher $F_1$-$score$ and 0.6\% higher $Sensitivity$ than VGAN. Several zoomed in images are displayed in Fig. \ref{fig:1} which indicate that CSAU obtain good vessel structures on STARE either.

On HRF, CSAU is compared with Odstrcilik, Vostatek(Soares), Vostatek(Sofka) and  Orlando. The results of different methods are differed a lot on general segmentation metrics. Thus we did not compute the metrics of $ACC_{cs}$ for further analysis. CSAU gets the highest scores and values in this group of experiments. Compared with Orlando, the $F_1$-$score$ is enhanced by more than 14 percent.

\begin{table}
\center
\caption{Comparison of different methods on STARE.}
\scalebox{0.75}{
\setlength{\tabcolsep}{0.7mm}{
\begin{tabular}{ccccccc}
\hline
\multirow{2}{*}{Methods} & \multicolumn{6}{c}{STARE} \\
\cline{2-7} & ROC AUC & PR AUC & $F_1$-$score$ & $Sensitivity$ & $Accuracy$ & $ACC_{cs}$ \\
\hline
HED\cite{xie2015holistically} & 0.9764 & 0.8888 & 0.8057 & 0.8200 & 0.9588 & 0.7257 \\
\hline
Wavelets\cite{soares2006retinal} & 0.9694 & 0.8433 & 0.7756 & 0.7817 & 0.9529 & 0.7226 \\
\hline
DRIU\cite{maninis2016deep} & 0.9772 & 0.9101 & 0.8323 & {\color{blue}0.8380} & 0.9648 & 0.7667 \\
\hline
VGAN\cite{son2017retinal} & {\color{blue}0.9777} & {\color{blue}0.9159} & {\color{blue}0.8353} & 0.8350 & {\color{blue}0.9657} & {\color{blue}0.7694} \\
\hline
CSAU & {\color{red} 0.9834} & {\color{red}0.9206} & {\color{red}0.8435} & {\color{red}0.8465} & {\color{red}0.9673} & {\color{red}0.7878} \\
\hline
\end{tabular}}}
\end{table}


\begin{table}
\center
\caption{Comparison of different methods on HRF.}
\scalebox{0.8}{
\setlength{\tabcolsep}{0.9mm}{
\begin{tabular}{cccccc}
\hline
\multirow{2}{*}{Methods} & \multicolumn{5}{c}{HRF} \\
\cline{2-6} & \tabincell{c}{ROC AUC} & \tabincell{c}{PR AUC} & \tabincell{c}{$F_1$-$score$} & $Precision$ & $Sensitivity$ \\
\hline
Odstrcilik\cite{odstrcilik2013retinal} & 0.967 & -- & {\color{blue}0.7316} & 0.6950 & {\color{blue}0.7772} \\
\hline
\tabincell{c}{Vostatek\\ (Soares)\cite{vostatek2017performance}} & {\color{blue}0.97} & -- & -- & -- & 0.7340 \\
\hline
\tabincell{c}{Vostatek\\ (Sofka)\cite{vostatek2017performance}} & 0.937 & -- & -- & -- & 0.5830 \\
\hline
Orlando\cite{orlando2017retinal} & -- & -- & 0.7168 & {\color{blue}0.7199} & 0.7201 \\
\hline
CSAU & {\color{red}0.9867} & {\color{red}0.9047} & {\color{red}0.8171} & {\color{red}0.8043} & {\color{red}0.8303} \\
\hline
\end{tabular}}}
\end{table}

\subsection{Experiment Analysis}
\label{sec:Experiment Analysis}
\begin{figure}
\centering
\includegraphics[width=0.7\linewidth]{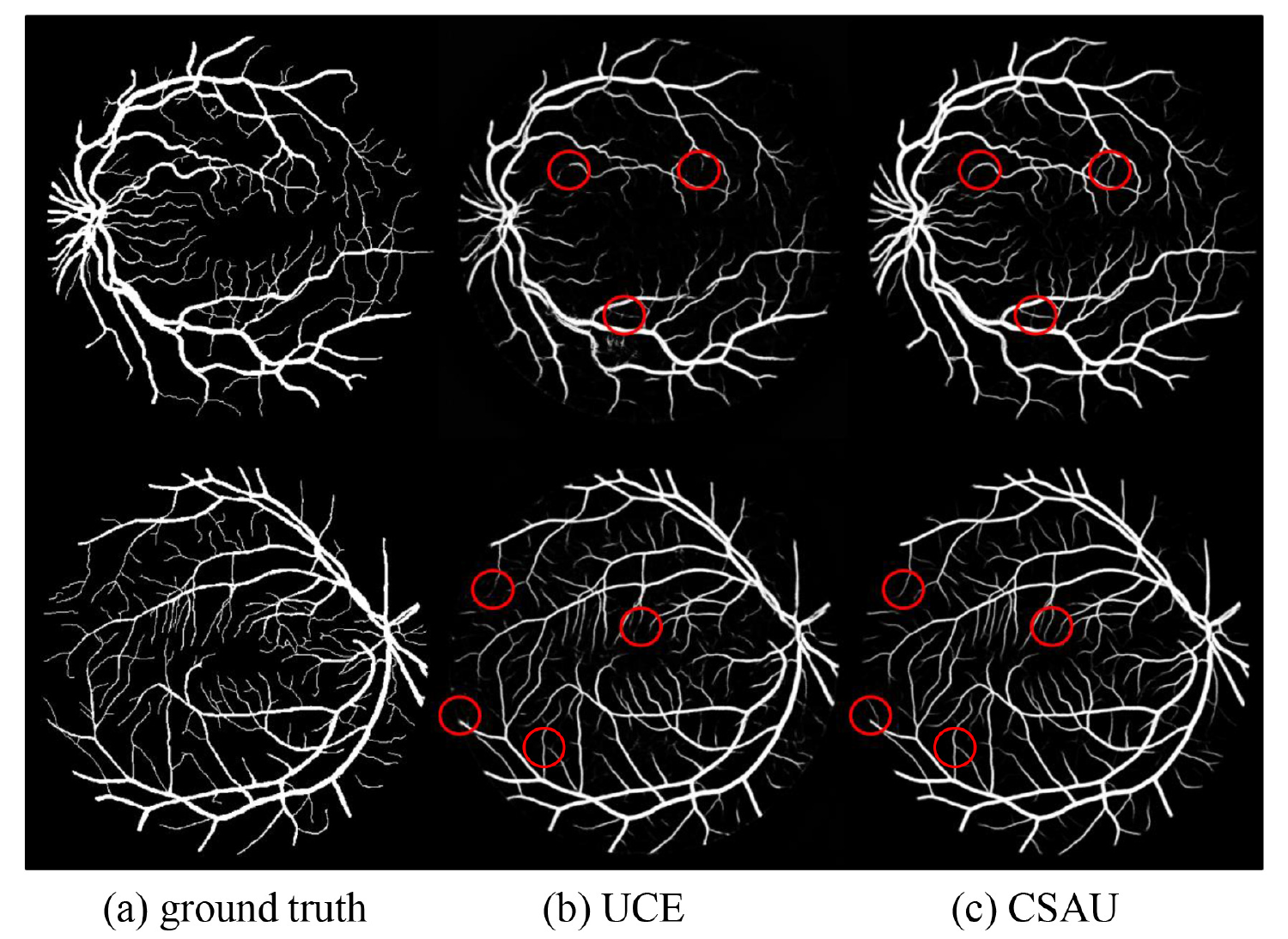}
\caption{Comparison between UCE and CSAU.}
\label{fig:12}
\end{figure}

To explore the reason why CSAU could get good performance, we carried out extra experiments on the datasets. We tried four different combinations. They are U-Net with CE loss(UCE), U-Net with CS loss(UCS), Attention U-Net with CE loss (AUCE) and Attention U-Net with CS loss(CSAU). Table 6 and 7 display the results of different combinations on DRIVE and STARE respectively. The table of HRF are provided in supplementary materials. From the results, we could find that either the usage of the proposed attention mechanism or that of the CS loss improves the performance. With both techniques, CSAU gets the best results in the group. Fig. \ref{fig:12} visually compares UCE and CSAU on an image of DRIVE. It is obvious that the proposed CSAU segments fine vessels more correctly while preserve topology structures well.

For quantitative analysis, on DRIVE, the result of CSAU is 0.6\% higher in $F_1$-$score$, 0.2\% higher in ROC AUC and 0.6\% higher in $Sensitivity$ than that of the UCE. As previously discussed, results on general metrics are not improved a lot. But in Fig. \ref{fig:12}, the enhancement is noticeble. To further analyze the source of contributions, we calculates the $ACC_{cs}$ on the results by different combinations. It could be seen that CSAU enhances the accuracy of segmentation mainly by improving the performance on boundaries and thin vessels. The other groups of experiments on STARE and HRF conform the effectiveness of the proposed method. Full experimental results could be found in the supplement materials.

\begin{table}
\center
\caption{Comparison of different combinations on DRIVE.}
\scalebox{0.75}{
\setlength{\tabcolsep}{0.7mm}{
\begin{tabular}{cccccccc}
\hline
\multirow{2}{*}{Methods} & \multicolumn{6}{c}{DRIVE} \\
\cline{2-7} & \tabincell{c}{$F_1$-$score$} & \tabincell{c}{PR AUC} & \tabincell{c}{ROC AUC} & $Sensitivity$ & $Accuracy$ & $ACC_{cs}$ \\
\hline
\tabincell{c}{UCE} & 0.8243 & 0.9084 & 0.9776 & {\color{blue}0.8318} & 0.9549 & 0.7435 \\
\hline
\tabincell{c}{UCS}  & 0.8255 & 0.9101 & {\color{blue}0.9802} & 0.8307 & {\color{blue}0.9554} & 0.7523 \\
\hline
\tabincell{c}{AUCE} & {\color{blue}0.8258} & {\color{blue}0.9111} & 0.9777 & 0.8303 & 0.9553 & {\color{blue}0.7524} \\
\hline
\tabincell{c}{CSAU} & {\color{red}0.8294} & {\color{red}0.9157} & {\color{red}0.9807} & {\color{red}0.8349}  & {\color{red}0.9563} & {\color{red}0.7751} \\
\hline
\end{tabular}}}
\end{table}

\begin{table}
\center
\caption{Comparison of different combinations on STARE.}
\scalebox{0.75}{
\setlength{\tabcolsep}{0.7mm}{
\begin{tabular}{cccccccc}
\hline
\multirow{2}{*}{Methods} & \multicolumn{6}{c}{STARE} \\
\cline{2-7} & $F_1$-$score$ & PR AUC & ROC AUC & $Sensitivity$ & $Accuracy$ & $ACC_{cs}$ \\
\hline
\tabincell{c}{UCE} & 0.8310 & 0.9096 & 0.9789 & 0.8350 & 0.9646 & 0.7513 \\
\hline
\tabincell{c}{UCS}\ & 0.8372 & 0.9155 & 0.9796 & {\color{red}0.8492} & 0.9656 & {\color{blue}0.7702} \\
\hline
\tabincell{c}{AUCE} & {\color{blue}0.8393} & {\color{blue}0.9202} & {\color{red}0.9842} & 0.8455 & {\color{blue}0.9663} & 0.7619 \\
\hline
\tabincell{c}{CSAU} & {\color{red}0.8435} & {\color{red}0.9206} & {\color{blue}0.9834} & {\color{blue}0.8465} & {\color{red}0.9673} & {\color{red}0.7878} \\
\hline
\end{tabular}}}
\end{table}

\section{Conclusions}
\label{sec:Conclusions}
In this paper, we proposed a very elegant symmetric neural network named connection sensitive attention U-Net for retinal vessels segmentation. Differed with other end-to-end semantic segmentation networks, the proposed CSAU not only concerned with pixel-level accuracy but also took care of topology structures by designing a novel connection sensitive loss and a new attention gate. The network was also learnt attention weights and concatenated it at the end of the network, which further improves the accuracy.

We verify the validity of CSAU on three public datasets: DRIVE, STARE, and HRF. The CSAU not only gets the highest $F_1$-$score$, ROC AUC and $Sensitivity$ on all the three datasets, but also performs well to segment the thin vessel structures, compared with the state-of-the-art methods. We also propose a new metrics named connection sensitive accuracy to evaluate the improvement on thin vessels segmentation. Based on it, we conclude that CSAU could segment thin vessels with high accuracy which is important for clinical diagnosis.

In the future, we will intend to try multiscale techniques and semi-supervised learning techniques to further enhance accuracy and efficiency.

\bibliographystyle{ieee}
\bibliography{references}
\end{document}